\def\aaaianonymous{true}
\documentclass[letterpaper]{article} 

\ifdefined\aaaianonymous
    \usepackage{aaai2026}  
\else
    \usepackage{aaai2026}              
\fi

\usepackage{times}  
\usepackage{helvet}  
\usepackage{courier}  
\usepackage[hyphens]{url}  
\usepackage{graphicx} 
\usepackage{natbib}  
\usepackage{caption} 
\newcommand{\name}[0]{DSC-Track}
\usepackage{algorithm}
\usepackage{algorithmic}
\usepackage{amsmath}
\usepackage{bm}
\usepackage{amssymb}
\usepackage{pifont}
\usepackage{booktabs}
\usepackage{multirow}

\usepackage{newfloat}
\usepackage{listings}
\DeclareCaptionStyle{ruled}{labelfont=normalfont,labelsep=colon,strut=off} 
\floatstyle{ruled}
\newfloat{listing}{tb}{lst}{}
\floatname{listing}{Listing}

\ifdefined\aaaianonymous
    \title{Delving into Dynamic Scene Cue-Consistency for 
Robust 3D Multi-Object Tracking}
\else
    \title{Delving into Dynamic Scene Cue-Consistency for 
Robust 3D Multi-Object Tracking}
\fi

\author{
    Haonan Zhang\textsuperscript{\rm 1,2}\thanks{Equal contribution. This work was done when Haonan was an intern at Fabu.},
    Xinyao Wang\textsuperscript{\rm 1,*},
    Boxi Wu\textsuperscript{\rm 1,3}\thanks{Corresponding authors.},
    Tu Zheng\textsuperscript{\rm 2,\dag},
    Wang Yunhua\textsuperscript{\rm 4},
    Zheng Yang \textsuperscript{\rm 2}
}

\affiliations{
    \textsuperscript{\rm 1}Zhejiang University, Hangzhou, China\\
    \textsuperscript{\rm 2}Fabu Inc., Hangzhou, China\\
    \textsuperscript{\rm 3}Daerwen AI, Hangzhou, China\\
    \textsuperscript{\rm 4}ShanDong Land-Sea-Nexus Digital Technology Co., Ltd., Shandong, China\\
}

\begin{document}

\maketitle

\begin{abstract}
3D multi-object tracking is a critical and challenging task in the field of autonomous driving. A common paradigm relies on modeling individual object motion, e.g., Kalman filters, to predict trajectories. While effective in simple scenarios, this approach often struggles in crowded environments or with inaccurate detections, as it overlooks the rich geometric relationships between objects. This highlights the need to leverage spatial cues. However, existing geometry-aware methods can be susceptible to interference from irrelevant objects, leading to ambiguous features and incorrect associations. To address this, we propose focusing on cue-consistency: identifying and matching stable spatial patterns over time. We introduce the Dynamic Scene Cue-Consistency Tracker (DSC-Track) to implement this principle. Firstly, we design a unified spatiotemporal encoder using Point Pair Features (PPF) to learn discriminative trajectory embeddings while suppressing interference. Secondly, our cue-consistency transformer module explicitly aligns consistent feature representations between historical tracks and current detections. Finally, a dynamic update mechanism preserves salient spatiotemporal information for stable online tracking. Extensive experiments on the nuScenes and Waymo Open Datasets validate the effectiveness and robustness of our approach. On the nuScenes benchmark, for instance, our method achieves state-of-the-art performance, reaching 73.2\% and 70.3\% AMOTA on the validation and test sets, respectively.  
Code Repository: \url{https://github.com/zhn12343333/DSCTrack}
\end{abstract}

\ifdefined\aaaianonymous
\else
\fi

\ifdefined\aaaianonymous

\section{Introduction}

Accurate and reliable 3D Multi-Object Tracking (MOT) is essential for ensuring the safety and reliability of autonomous driving systems. With the steady improvement of 3D object detectors~\cite{qi2018frustum, zhou2018voxelnet, lang2019pointpillars, yin2021center}, the tracking-by-detection paradigm~\cite{Zhang2021ByteTrackMT, Weng2020GNN3DMOT, Zeng2021Alphatrack} has remained a popular choice due to its ability to leverage rich appearance, motion, and geometric information of objects.

\begin{figure}[t]
\centering 
\includegraphics[width=\linewidth]{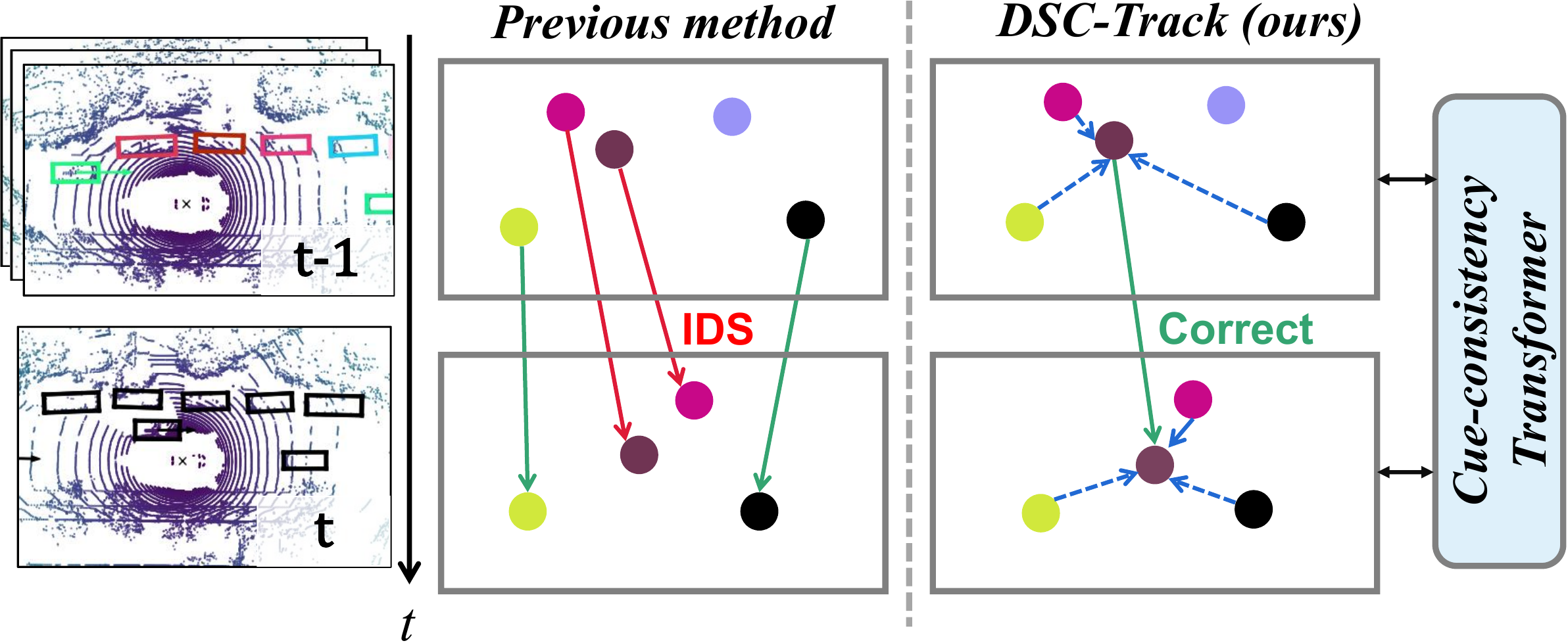}
\caption{Comparison of a conventional tracking method and our proposed DSC-Track. \textbf{Left}: Previous methods, such as those relying on individual object motion, are prone to ID Switches (IDS) when faced with ambiguous associations, like objects in close proximity. \textbf{Right}: Our method explicitly models the spatial cue-consistency among all targets using a Transformer, enabling robust and correct associations in challenging scenarios.}
\label{fig/fig1}
\end{figure}

One mainstream category of methods focuses on data association using predefined motion models~\cite{yin2021center,Weng20193DMT} (see Figure~\ref{fig/fig1}(Left)). Although these methods have achieved significant progress, their performance remains limited in complex scenarios due to their difficulty in effectively capturing multi-frame temporal information. To address this limitation, another line of work~\cite{kim2022polarmot,chu2023transmot} enhances feature representations by fusing geometric information with spatial dependencies of objects. However, this approach introduces new challenges: on one hand, its performance is highly susceptible to inaccuracies from upstream detectors; on the other hand, the indiscriminate utilization of spatial neighborhood nodes leads to temporally inconsistent feature representations, resulting in erroneous associations.

The performance bottleneck of the aforementioned methods lies in constructing robust features for tracklets, which raises two core challenges: 1) How to selectively leverage spatial cues to enhance feature discriminability? 2) How to ensure the consistency and stability of feature representations while incorporating multi-frame temporal information?
To tackle these challenges, we propose DSC-Track (see Figure~\ref{fig/fig1}(Right)), a method that deeply explores spatial-geometric cues in dynamic scenes to model consistent features between trajectories and detections. Its core idea is to leverage contextual information by searching for and associating a set of neighborhood nodes with spatio-temporal consistency for each trajectory in the historical space, rather than relying on prior knowledge (e.g., a fixed search radius). By actively suppressing interference from irrelevant objects and focusing on highly relevant spatial cues, our method significantly enhances the discriminability and stability of feature representations. This leads to more accurate data association and effectively mitigates the challenges posed by dynamic and crowded scenes.

Specifically, DSC-Track consists of the following steps:
~(1) A memory bank is constructed for each trajectory to store the historical information of the tracked object and the indices of spatially dependent nodes.
~(2) The unified spatiotemporal encoder aggregates unique feature embeddings of each trajectory from the historical dynamic space through Point Pair Features (PPF)~\cite{drost2010model} as spatial geometric representations, significantly reducing interference from irrelevant nodes.
~(3) A Transformer-based cue-consistency interaction module leverages spatial cues to extract consistent feature representations between trajectory embeddings and detections.
~(4) Feature-level affinity matrices constructed from consistency features enable data association, while the spatially dependent nodes in the memory bank are updated based on the latest dependencies of detection nodes mined by the cue-consistency interaction module, achieving robust object tracking in dynamic scenes.

To validate our approach, we conducted extensive experiments on two large-scale benchmarks: the nuScenes~\cite{Caesar2019nuScenesAM} and Waymo Open Datasets ~\cite{sun2020scalability}. Our ablation studies confirm the effectiveness of each proposed component in improving tracking performance and reducing heuristic dependencies. The key contributions of this work are summarized as follows:
\begin{itemize}
    \item We present \name{}, a novel transformer-based framework that leverages spatiotemporal cues for robust 3D multi-object tracking in dynamic scenes.
    \item A unified spatio-temporal aggregation module is proposed to capture consistent motion across trajectories from a spatio-temporal perspective, enabling the generation of highly discriminative features.
    \item A cue-consistent attention module is designed to mine consistent feature pairs from tracks and detections in dense scenes, significantly improving association accuracy.
    \item Using CenterPoint~\cite{yin2021center} detections as input, our method establishes a new state-of-the-art on the nuScenes benchmark, achieving 73.2\% and 70.3\% AMOTA on the validation and test sets, respectively, and demonstrates strong, generalizable performance on the Waymo dataset, confirming the robustness of our design.
\end{itemize}

\section{Related Work}
\else
\section{Introduction}
\fi

\paragraph{Tracking-by-Detection in MOT.}
The Tracking-by-Detection (TBD) paradigm dominates Multi-Object Tracking (MOT). Early 2D methods evolved from simple motion models like SORT~\cite{Bewley2016SimpleOA} to more advanced techniques incorporating appearance features and refined association strategies~\cite{Wojke2017SimpleOA, Zhang2021ByteTrackMT}. This philosophy extends to 3D MOT, where methods often pair high-fidelity detectors with similar association techniques~\cite{Weng20193DMT, zhang2023bytetrackv2}. More recent learning-based approaches utilize Graph Neural Networks (GNNs)~\cite{braso2020learning, zaech2022learnable} or Transformers~\cite{chu2023transmot, Ding2022End2End} to model relationships, yet many still primarily focus on individual object cues, overlooking rich structural information in the spatial context.

\paragraph{Spatial Information in MOT.}
Leveraging spatial information is crucial for robust tracking. While some methods model spatial relationships using graphs~\cite{he2021learnable, kim2022polarmot} or Transformers~\cite{chu2023transmot, Ding2023MOTFormer}, they often face a dilemma: either restricting interactions to local neighborhoods or indiscriminately aggregating global information, which introduces interference from irrelevant objects. In contrast, our method addresses this by adaptively searching for a unique set of spatially consistent dependent nodes for each track. This allows aggregation of highly relevant contextual cues while suppressing interference for more robust association.

\section{Approach}
\label{sec:approach}

\begin{figure*}[t]
\centering 
\includegraphics[width=\linewidth]{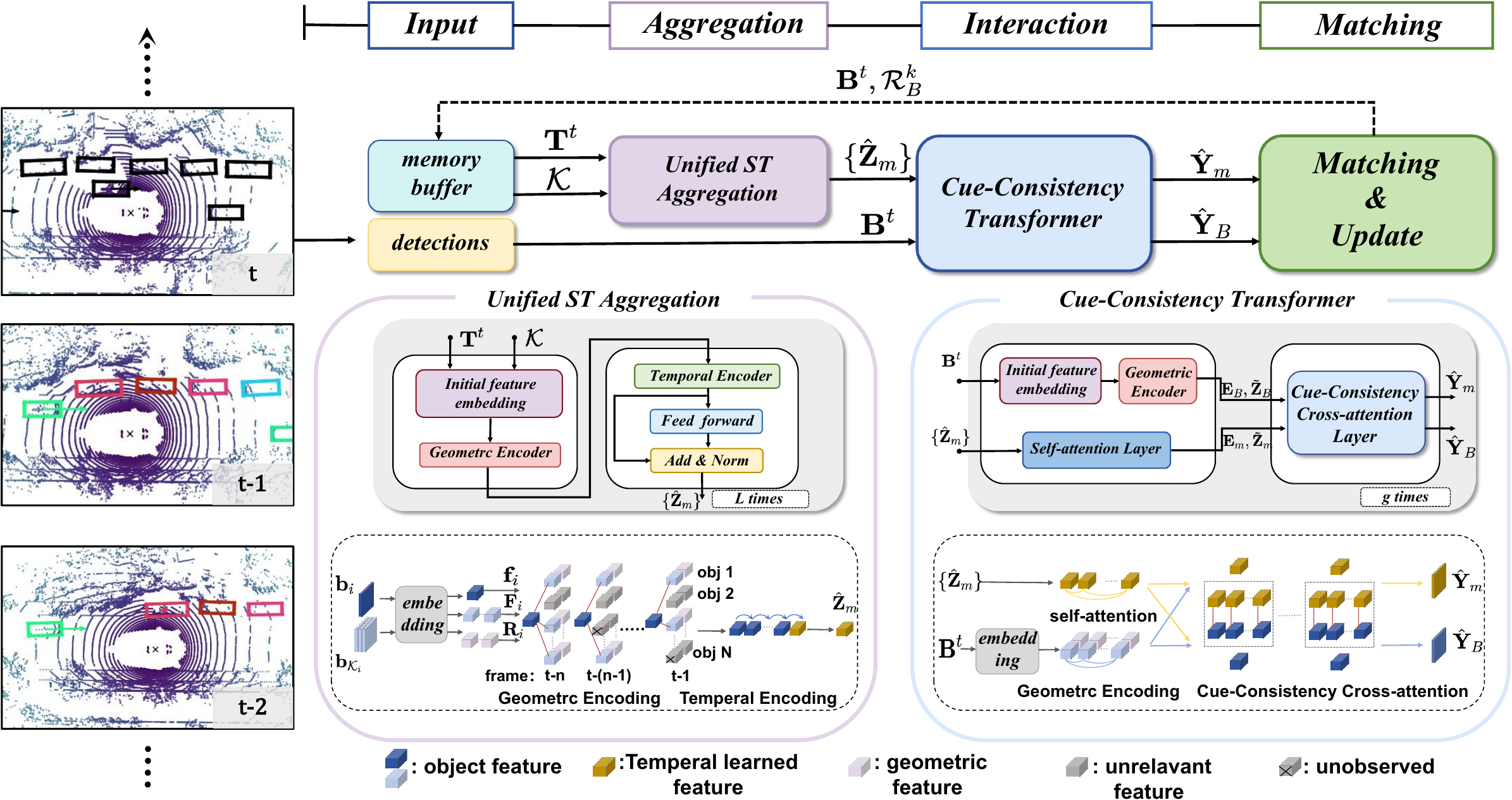}
\caption{
    \textbf{The overall architecture of our \name{} framework.} 
    Our model takes historical track information and new 3D detections as input. 
    (1) The \textbf{Unified Spatio-Temporal Aggregation} module first generates a discriminative feature representation, $\{\hat{\mathbf{Z}}_m\}$, for each track by leveraging its historical and spatial context. 
    (2) Then, the \textbf{Cue-Consistency Transformer} interacts these track features with detection features ($\mathbf{B}^t$) to mine consistent cues, yielding enhanced representations for both. 
    (3) Finally, in the \textbf{Matching and Update} stage, an affinity matrix is computed from these enhanced features for data association, and the memory buffer is updated for the next frame.
}
\label{fig:fig2}
\end{figure*}

\subsection{Preliminaries}

\paragraph{Problem Statement.}
Given a set of $N$ 3D bounding box candidates, $\mathbf{B}^t = \{\mathbf{b}_{j}^t \mid j = 1, \dots, N\}$, detected at the current frame $t$, and a set of $M$ active tracks, $\mathbf{T}^{t-1} = \{\tau_{i}^{t-1} \mid i = 1, \dots, M\}$, from the previous frame, our primary goal in 3D Multi-Object Tracking is to establish associations between the active tracks in $\mathbf{T}^{t-1}$ and the new detections in $\mathbf{B}^t$. These inputs are typically provided by a 3D object detector, such as CenterPoint~\cite{yin2021center}.

\paragraph{Detection Node Representation.}
Each detection candidate $\mathbf{b}_{i}$ in $\mathbf{B}^t$ is represented by a state vector, following conventions from prior works~\cite{zaech2022learnable,Ding2023MOTFormer}. Specifically, for the nuScenes dataset, this state is defined as $\mathbf{b}_{i}=[p_i,\theta_i,s_i,v_i,c_i,o_i] \in \mathbb{R}^{17}$. This vector encapsulates the 3D center position $p_i \in \mathbb{R}^3$, heading angle $\theta_i \in \mathbb{R}$, 3D box size $s_i \in \mathbb{R}^3$, velocity $v_i \in \mathbb{R}^2$, a one-hot encoded class label $c_i \in \mathbb{R}^7$, and the detection confidence score $o_i \in \mathbb{R}$.

\paragraph{Trajectory Representation.}
To maintain temporal context, we represent each active track $\tau_i$ using a memory bank. This bank stores two key components: its historical state information, $\mathbf{M}_i$, and the IDs of its spatially dependent neighboring tracks, $\mathcal{K}_i$. Formally, a track is defined as $\tau_{i} = \{\mathbf{M}_i, \mathcal{K}_i\}$. Here, $\mathbf{M}_i \in \mathbb{R}^{T_{\text{max}} \times 17}$ is a matrix containing the state vectors of the track over the past $T_{\text{max}}$ time steps. $\mathcal{K}_i$ denotes the set of tracking identifiers for its spatial neighbors. After the association step in each frame, both $\mathbf{M}_i$ and $\mathcal{K}_i$ are updated to reflect the latest state and spatial relationships, with the update mechanism detailed in Section \ref{sec:node_update}.


\subsection{Unified Spatiotemporal Aggregation}
\label{sec:backbone}
\paragraph{Overview.}
To generate highly discriminative features for each of the $M$ active tracks in $\mathbf{T}^{t-1}$, we propose a unified spatio-temporal aggregation module. The core idea is to leverage spatially consistent geometric relationships across frames to produce robust track representations, denoted as $\{\hat{\mathbf{Z}}_m\} \in \mathbb{R}^{M \times d}$. As shown in Figure~\ref{fig:fig2}, the architecture of this module consists of a Geometric Encoder followed by a Temporal Encoder. This two-stage process is iterated $L$ times to progressively refine the features.

\begin{figure}[]
\centering 
\includegraphics[width=0.48\textwidth]{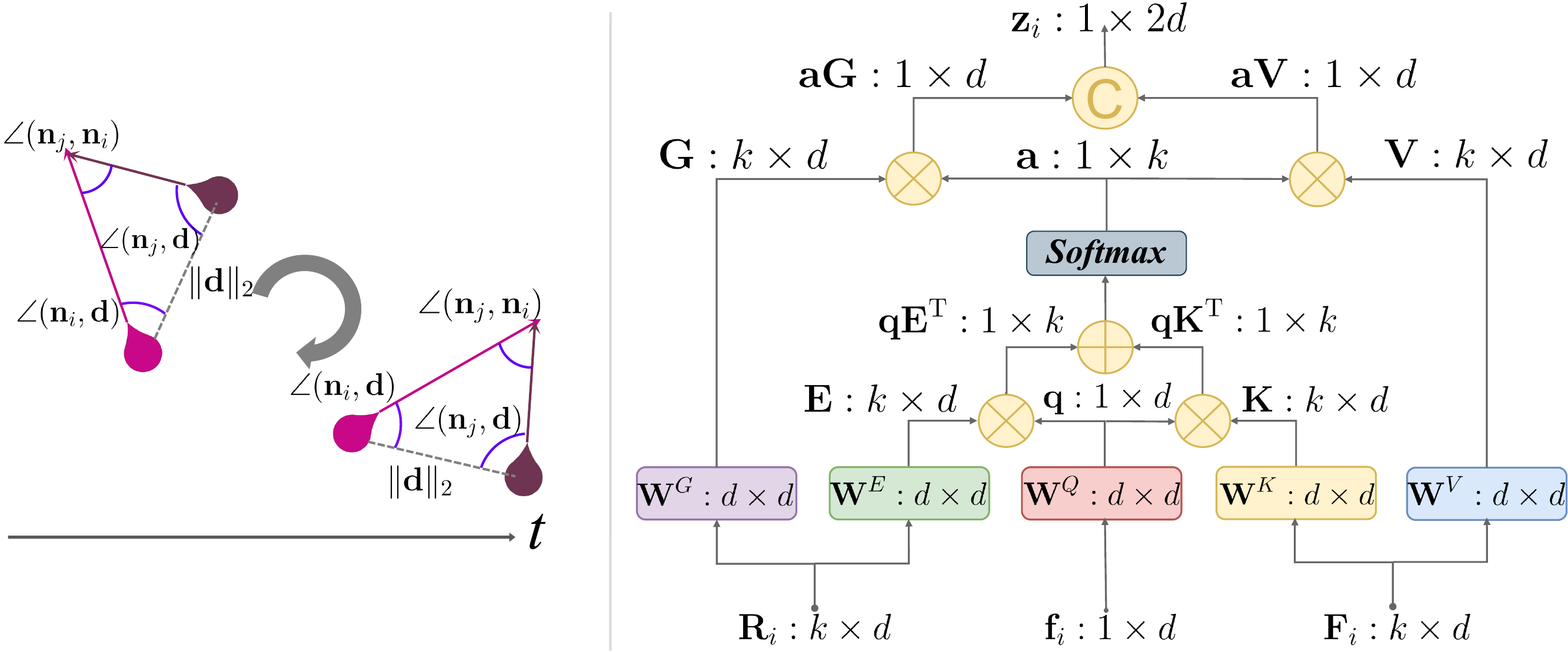}
\caption{\textbf{Left:} Illustration of the rotation-invariance of our Point Pair Feature (PPF). \textbf{Right:} Details of the Geometric Inject Attention (GIA) module used for aggregation.}
\label{fig/fig3}
\end{figure}

\paragraph{Initial Feature Embedding.}
\label{sec:ppf_encoder}
Modeling geometric relationships from absolute object states is unreliable, as these features lack rotation-invariance and become unstable when objects maneuver (e.g., a vehicle turning). To address this, we construct a local geometric embedding using the Point Pair Feature (PPF)~\cite{drost2010model}. By capturing the relative geometry—defined by the distance and inter-object angles (see Figure~\ref{fig/fig3}(Left))—PPF provides a robust representation that remains consistent across rotations and viewpoint changes, ensuring stable geometric modeling.

For a given reference object $\mathbf{b}_{i}$ and its set of $k$ spatially-dependent neighbors $\{\mathbf{b}_{j}\}_{j \in \mathcal{K}_i}$, we first extract their positions $p$ and heading angles $\theta$. Each heading angle is converted to a 2D direction vector $\mathbf{n}$ from the Bird's-Eye View (BEV). The PPF encodes the relationship between the reference object $\mathbf{b}_i$ and each neighbor $\mathbf{b}_{j}$ as a 4D vector:
\begin{equation}
\mathbf{e}_{i,j} = \left( \| \mathbf{d} \|_2, \angle (\mathbf{n}_i, \mathbf{d}), \angle (\mathbf{n}_j, \mathbf{d}), \angle (\mathbf{n}_j, \mathbf{n}_i) \right) \in \mathbb{R}^{4},
\label{eq:ppf_feature}
\end{equation}
where $\mathbf{d} = p_j - p_i$, and $\angle (\cdot, \cdot)$ computes the angle between two vectors. By concatenating these features for all $k$ neighbors, we form a local geometric matrix $\mathbf{E}_i \in \mathbb{R}^{k \times 4}$.

This geometric matrix is then processed to create our final feature triplet. First, $\mathbf{E}_i$ is enhanced with sinusoidal positional encodings~\cite{Vaswani2017attention} and projected by an MLP to form the relative geometry embedding $\mathbf{R}_i \in \mathbb{R}^{k \times d}$. Concurrently, the state vectors of the reference object $\mathbf{b}_{i}$ and its neighbors $\{\mathbf{b}_{j}\}_{j \in \mathcal{K}_i}$ are projected by a shared MLP into contextual embeddings: the reference feature $\mathbf{f}_{i}\in \mathbb{R}^{1 \times d}$ and neighbor features $\mathbf{F}_{i} \in \mathbb{R}^{k \times d}$.

These three components form the geometric triplet $\mathcal{G}_i = (\mathbf{f}_{i}, \mathbf{F}_{i}, \mathbf{R}_i)$, which serves as the rich, geometrically-aware input for the subsequent encoder.

\paragraph{Geometric Encoder.}
The geometric encoder processes the geometric triplets frame-by-frame for each track. For a given triplet $\mathcal{G}_i^t$ at frame $t$, the encoder aggregates the contextual features of neighbors ($\mathbf{F}_{i}$) onto the reference feature ($\mathbf{f}_{i}$), guided by their rich geometric relationships ($\mathbf{R}_{i}$). This produces an updated, geometrically-aware feature $\mathbf{z}_i^t \in \mathbb{R}^{d}$. For simplicity, we omit the frame index $t$ in the following description.

The core of this process is our \textbf{Geometric Inject Attention (GIA)} module, inspired by~\cite{yu2023rotation} and depicted in Figure~\ref{fig/fig3}~(right). The GIA module enhances standard attention by injecting geometric information into both the attention score calculation and the value aggregation. The process involves several key steps. \textbf{First,} we project the input triplet into components for the attention mechanism. The contextual embeddings are projected into a standard query, key, and value:
\begin{equation}
\mathbf{q} = \mathbf{f}_i \mathbf{W}^q, \quad 
\mathbf{K} = \mathbf{F}_{i} \mathbf{W}^K, \quad 
\mathbf{V} = \mathbf{F}_{i} \mathbf{W}^V,
\end{equation}
where $\mathbf{W}^q, \mathbf{W}^K, \mathbf{W}^V \in \mathbb{R}^{d \times d}$ are learnable matrices. Concurrently, the relative geometry embedding $\mathbf{R}_{i}$ is projected into a geometric bias for attention scores, and geometric features for value aggregation:
\begin{equation}
\mathbf{E} = \mathbf{R}_{i} \mathbf{W}^E, \quad
\mathbf{G} = \mathbf{R}_{i} \mathbf{W}^G.
\end{equation}
\textbf{Second,} the attention scores are computed by injecting the geometric bias $\mathbf{E}$ into the standard dot-product attention:
\begin{equation}
\mathbf{a} = \text{Softmax} \left( \frac{\mathbf{q} \mathbf{K}^T + \mathbf{q} \mathbf{E}^T}{\sqrt{d}} \right).
\end{equation}
\textbf{Finally,} the output $\mathbf{z}_i$ is obtained by using the attention scores $\mathbf{a}$ to aggregate both the contextual values $\mathbf{V}$ and the geometric features $\mathbf{G}$. This is followed by a standard residual connection and a Feed-Forward Network (FFN):
\begin{align}
\mathbf{z}_i' &= \mathbf{f}_i + \text{MLP}_{\text{GIA}}\left( \text{concat}(\mathbf{a} \mathbf{V}, \mathbf{a} \mathbf{G}) \right), \label{eq:gia_agg} \\
\mathbf{z}_i &= \text{FFN}(\text{LN}(\mathbf{z}_i')), \label{eq:geo_output}
\end{align}
where $\text{LN}(\cdot)$ is Layer Normalization~\cite{Ba2016LayerN} and $\text{FFN}$ consists of a two-layer MLP with a residual connection.

\paragraph{Analysis.}
By explicitly fusing geometric embeddings with contextual information, our method enhances feature representations. At the \textbf{intra-frame} level, each trajectory leverages unique spatial dependencies to obtain a highly distinctive feature. This, in turn, enhances robustness at the \textbf{inter-frame} level: by producing stable and geometrically-aware features for each frame, this encoder provides a superior input to the subsequent temporal encoder, which can then more effectively suppress noise and ensure stable spatial relationships over time.

\paragraph{Temporal Encoder.}
After the Geometric Encoder, we obtain a sequence of frame-wise feature representations for each track, denoted as $\mathbf{Z}_i = [\mathbf{z}_i^{t-T_{\text{max}}+1}, \dots, \mathbf{z}_i^{t}] \in \mathbb{R}^{T_{\text{max}} \times d}$. The role of the Transformer-based Temporal Encoder is to aggregate this historical sequence and reason about the track's latent state.

Inspired by BERT~\cite{devlin2018bert, huang2023delving}, we introduce a learnable \textbf{track token}, $\mathbf{z}_m \in \mathbb{R}^{1 \times d}$, which acts as a summary for the entire trajectory. This token is prepended to the historical sequence, and the combined input is fed into a self-attention module. The query, key, and value matrices are computed as follows:
\begin{align}
\begin{split}
\mathbf{Q} &= [\mathbf{z}_m, \mathbf{Z}_i]\mathbf{W}_T^Q, \\
\mathbf{K} &= [\mathbf{z}_m, \mathbf{Z}_i]\mathbf{W}_T^K, \\
\mathbf{V} &= [\mathbf{z}_m, \mathbf{Z}_i]\mathbf{W}_T^V,
\end{split} \label{eq:QKV_temporal}
\end{align}
where $\mathbf{W}_T^*$ are the learnable projection matrices and $[\cdot,\cdot]$ denotes concatenation along the sequence dimension. To prevent information leakage from the future, a causal mask is applied during the self-attention computation.

The full temporal encoder layer then processes the attention output with a feed-forward network (FFN), including residual connections and layer normalization:
\begin{equation}
[\hat{\mathbf{z}}_m, \hat{\mathbf{Z}}_i] = \mathrm{FFN}(\mathrm{SelfAttn}(\mathbf{Q},\mathbf{K},\mathbf{V})),
\label{eq:temporal_final}
\end{equation}
The updated track token, $\hat{\mathbf{z}}_m$, ultimately encapsulates the rich, aggregated spatio-temporal representation of the trajectory, providing a robust feature for the subsequent interaction modules.

\subsection{Cue-Consistent Attention Module}
\label{sec:interaction}

\paragraph{Overview.}
Given the aggregated track features $\{\hat{\mathbf{z}}_m\}$ and new detections, our goal is to robustly associate them by mining their underlying feature consistency. To achieve this, we design a Cue-Consistency Attention Module. As illustrated in Figure~\ref{fig:fig2}, this module first enhances both track and detection features independently and then interacts them to produce final, matchable representations. This process can be iterated $g$ times to progressively refine the matching cues.

\paragraph{Self-Information Encoder.}
First, we independently enhance the features of tracks and detections. For the $M$ input track features, aggregated in a matrix $\hat{\mathbf{Z}}_m \in \mathbb{R}^{M \times d}$, we use a vanilla self-attention layer~\cite{Vaswani2017attention} to mine intra-feature relationships, producing globally-aware features $\tilde{\mathbf{Z}}_{m} \in \mathbb{R}^{M \times d}$.

Concurrently, we process the new detections. For each detection $\mathbf{b}_{i} \in \mathbf{B}^t$, we apply our geometric encoder (denoted as $\delta$) to aggregate information from its neighborhood. Note that unlike the selective sampling in Section~\ref{sec:ppf_encoder}, here we use \textbf{all} other objects as neighbors to form the geometric triplet. This ensures that we do not miss potentially important information, given that detector outputs can be unstable. This process yields the initial detection features $\mathbf{Z}_B = [\delta(\mathbf{b}_{1}), \dots, \delta(\mathbf{b}_{N})] \in \mathbb{R}^{N \times d}$. These are then passed through another self-attention layer to yield enhanced features $\tilde{\mathbf{Z}}_B \in \mathbb{R}^{N \times d}$.

Crucially, during these two self-attention steps, we cache the resulting attention score matrices, $\mathbf{E}_{m} \in \mathbb{R}^{M \times M}$ and $\mathbf{E}_{B} \in \mathbb{R}^{N \times N}$, for subsequent use in our cue-consistent cross-attention.

\paragraph{Cue-Consistent Cross-Attention.}
Cross-attention is a typical module for exchanging information between two feature sets~\cite{huang2021predator,wang2019deep,sun2021loftr}. Inspired by~\cite{chen2024dynamic}, we design a \textit{cue-consistent cross-attention} module to model the consistency between $\tilde{\mathbf{Z}}_{m}$ and $\tilde{\mathbf{Z}}_B$. For clarity, we demonstrate the process for updating detection features by attending to tracks; the reverse process is symmetric.

The procedure unfolds in three steps. \textbf{First,} to expand the structural information for each feature, we extract "cues" guided by the pre-computed self-attention matrices. For each detection $i$, its cue is constructed from the features of its top-$k$ most similar peer detections, identified via $\mathbf{E}_B$:
\begin{equation}
\mathbf{C}_{B,i} = \text{Gather}(\tilde{\mathbf{Z}}_B, \text{topk}(\mathbf{E}_B(i, \cdot), k)) \in \mathbb{R}^{k \times d}.
\label{eq:cue_extraction}
\end{equation}
Here, $\text{topk}(\cdot, k)$ returns the indices of the $k$ largest values, and $\text{Gather}(\cdot)$ collects the features at these indices. Similarly, for each track $j$, we extract its cue $\mathbf{C}_{m,j} \in \mathbb{R}^{k \times d}$ using $\tilde{\mathbf{Z}}_m$ and $\mathbf{E}_m$.

\textbf{Second,} we compute the attention from a detection $i$ to all tracks $j=1, \dots, M$. The attention weight is not based on direct feature similarity, but on the consistency score $s_{ij}$ between their respective cues, $\mathbf{C}_{B,i}$ and $\mathbf{C}_{m,j}$. We model this score by performing a feature-wise dot-product between the two cue matrices and summing the result:
\begin{equation}
s_{ij} = \sum_{l=1}^{k} \frac{ (\mathbf{C}_{B,i}[l] \mathbf{W}^Q) (\mathbf{C}_{m,j}[l] \mathbf{W}^K)^T }{\sqrt{d}},
\end{equation}
where $\mathbf{C}[l]$ denotes the $l$-th feature vector in the cue matrix, and $\mathbf{W}^Q, \mathbf{W}^K$ are learnable projection matrices. This process learns to measure the consistency confidence between the two neighborhood structures.

\textbf{Finally,} the consistency scores $\{s_{ij}\}_{j=1}^M$ are normalized using a softmax to produce the attention weights. The updated detection feature $\mathbf{y}_{B,i}$ is then the weighted average of all track features:
\begin{equation}
\mathbf{y}_{B,i} = \sum_{j=1}^{M} \text{softmax}_j(s_{ij}) \cdot (\tilde{\mathbf{z}}_{m,j} \mathbf{W}^V).
\end{equation}
The full matrix of updated features $\mathbf{Y}_{B}$ is further sent into a feed-forward layer to obtain the enhanced message $\hat{\mathbf{Y}}_B$. The cue-consistent cross-attention is also applied in the reverse direction, yielding $\hat{\mathbf{Y}}_m$.

\paragraph{Computational Cost Reduction.}
It is important to note that computing cue-consistent cross-attention for all pairs is computationally demanding. To reduce this burden, we utilize prior knowledge from online MOT. We first predict the potential locations of historical trajectories in the current frame by applying an FFN: $p_m = \text{FFN}(\{\hat{\mathbf{Z}}_m\})$. Interactions are then established only between objects of the same class, and are further pruned based on a class-specific distance threshold. This threshold is calculated from dataset statistics of maximum speed per class, following~\cite{zaech2022learnable}.

\subsection{Feature Matching and Update}

\paragraph{Overview.}
After extracting the enhanced features $\hat{\mathbf{Y}}_{m}$ and $\hat{\mathbf{Y}}_{B}$, we first compute a feature-based affinity matrix to associate tracks with detections. Then, for each successful match, we update the track's stored neighborhood information for the next frame.

\paragraph{Feature-based Matching.}
To compute the affinity between all track-detection pairs, we expand the feature matrices $\hat{\mathbf{Y}}_{m} \in \mathbb{R}^{M \times C}$ and $\hat{\mathbf{Y}}_{B} \in \mathbb{R}^{N \times C}$ to a compatible shape of $\mathbb{R}^{M \times N \times C}$ by tiling. An MLP then processes their element-wise difference to predict a match score:
\begin{align}
\mathbf{A} = \text{Sigmoid}(\text{MLP}(\hat{\mathbf{Y}}_{m}^e - \hat{\mathbf{Y}}_{B}^e)),
\end{align}
where $\mathbf{A} \in \mathbb{R}^{M \times N}$ is the final affinity matrix, whose entries represent matching probabilities.

\paragraph{Neighborhood Update.}
\label{sec:node_update}
To keep the context of each track current, its neighborhood information must be updated over time. When a track $j$ is associated with a detection $i$, we update its stored set of neighbor IDs with those of detection $i$, which are defined by the cue $\mathbf{C}_{B,i}$ from Equation~\ref{eq:cue_extraction}.

\paragraph{Loss Function.}
The model is trained with a composite loss. For association, we use the Focal Loss~\cite{Lin2017FocalLF} ($\mathcal{L}_a$) with $\alpha=-1, \gamma=1$. For position regression, we use a smooth $\ell_1$ loss ($\mathcal{L}_p$). The total loss is:
\begin{equation}
\mathcal{L} = \mathcal{L}_a + \lambda_p \mathcal{L}_p,
\end{equation}
where the balancing weight $\lambda_p$ is set to $0.5$.

\section{Experiment}
\subsection{Experiment Setup}
\label{sec:exp_setup}

\begin{table*}[]
\centering
\setlength{\tabcolsep}{0.8mm}
\begin{tabular}{lc|cccc|ccccccc} 
\toprule
\multirow{2}{*}{Method} & \multirow{2}{*}{Additional Cues} & \multicolumn{4}{c}{Average Metrics} & \multicolumn{7}{c}{Class-specific AMOTA} \\ 
\cmidrule(lr){3-6} \cmidrule(lr){7-13} 
& & AMOTA$\uparrow$ & AMOTP$\downarrow$ & MOTA$\uparrow$  & IDS$\downarrow$ & car & ped & bicycle & bus & motor & trailer & truck \\
\midrule

CenterPoint  & -- & 0.638 & 0.555 & 0.537 & 760 & 0.829 & 0.767 & 0.321 & 0.711 & 0.591 & 0.651 & 0.599 \\ 
CBMOT   & -- & 0.649 & 0.592 & 0.545 & 557 & 0.828 & 0.794 & 0.372 & 0.704 & 0.592 & 0.667 & 0.587 \\ 
SimpleTrack$^\ddagger$   & -- & 0.668 & 0.550 & 0.566 & 575 & 0.823 & 0.796 & 0.407 & 0.715 & 0.674 & 0.673 & 0.587 \\ 
ImmortalTracker  & -- & 0.677 & 0.599 & 0.572 & 320 & 0.833 & 0.816 & \textbf{0.416} & 0.716 & 0.689 & 0.675 & 0.596 \\ 
PolarMOT-offline$^\dagger$  & -- & 0.664 & 0.566 & 0.561 & \textbf{242} & 0.853 & 0.806 & 0.349 & 0.708 & 0.656 & 0.673 & 0.602 \\ 
3DMOTFormer  & -- & 0.682 & 0.496 & 0.556 & 438 & 0.821 & 0.807 & 0.374 & 0.749 & \textbf{0.705} & 0.696 & 0.626 \\ 
\midrule
NEBP  & 3D appearance & 0.683 & 0.624 & 0.584 & 227 & 0.835 & 0.802 & 0.447 & 0.708 & 0.698 & 0.69 & 0.598 \\ 
ShaSTA  & 3D appearance & 0.696 & 0.540 & 0.578 & 473 & 0.838 & 0.81 & 0.41 & 0.733 & 0.727 & 0.704 & 0.65 \\ 
\midrule
\textbf{DSC-Track(ours)} & -- & \textbf{0.703} & \textbf{0.476} & \textbf{0.575} & 301 & \textbf{0.858} & \textbf{0.830} & 0.401 & \textbf{0.755} & 0.703 & \textbf{0.719} & \textbf{0.653} \\
\bottomrule
\end{tabular}
\caption{Results on nuScenes test set using CenterPoint detections. $^\dagger$denotes offline methods, $^\ddagger$denotes using 10Hz data. The best results are highlighted in bold}
\label{tab:nuscenes_test}
\end{table*}

\begin{table}[]
\centering
\setlength{\tabcolsep}{1mm}
\begin{tabular}{c|ccccccc}
\toprule
Method & AMOTA$\uparrow$ & AMOTP$\downarrow$ & MOTA$\uparrow$  & IDS$\downarrow$ \\
\midrule
CenterPoint & 0.665 & 0.567 & 0.562 & 562 \\
SimpleTrack & 0.696 & 0.547 & 0.602 & -- \\
ImmortalTracker & 0.702 & -- & 0.601 & -- \\
PolarMOT-offline & 0.711 & -- & -- & \textbf{213} \\
3DMOTFormer & 0.712 & 0.515 & 0.607 & 341 \\
\midrule
\textbf{DSC-Track(ours)} & \textbf{0.732} & \textbf{0.498} & \textbf{0.625} & 298\\
\bottomrule
\end{tabular}
\caption{Comparison with state-of-the-art tracking-by-detection approaches on the nuScenes validation split.}
\label{tab:nuscenes_val}
\end{table}

\paragraph{Dataset.}
We conduct experiments on two large-scale autonomous driving benchmarks. The Waymo Open Dataset~\cite{sun2020scalability} provides 1150 sequences (798 training, 202 validation, 150 testing) of 20-second driving data with 3D labels for three classes: Vehicle, Pedestrian, and Cyclist. The nuScenes dataset~\cite{Caesar2019nuScenesAM}, our primary benchmark, comprises 1000 sequences of similar length. It is distinguished by a richer sensor suite (32-beam LiDAR, RADAR, six cameras) and provides 3D tracking annotations at 2Hz for seven object classes, making it ideal for leveraging multi-modal data.

\paragraph{Metrics.}
We follow the official nuScenes tracking benchmark protocol for evaluation. The primary metrics are Average Multi-Object Tracking Accuracy (AMOTA) and Average Multi-Object Tracking Precision (AMOTP)~\cite{Weng20193DMT}. We also report secondary metrics from CLEAR MOT~\cite{Bernardin2008EvaluatingMO}, including Multi-Object Tracking Accuracy (MOTA), ID Switches (IDS).

\paragraph{Implementation details.}
We construct the training dataset using outputs from CenterPoint~\cite{yin2021center}. Following 3DMOTFormer~\cite{Ding2023MOTFormer}, we trained our tracker for 16 epochs on randomly sampled mini-sequences of length $T$=10. For the memory buffer, we set the maximum temporal length $T_{max}$=7 and select the top $k=3$ spatially dependent objects to capture the historical information for each tracklet. In the cue-consistent attention module, we set top $k=3$ to perform the cue-consistent cross attention. For optimization, we use the AdamW~\cite{loshchilov2017decoupled} optimizer with an initial learning rate of $2e^{-4}$ and a cosine decay schedule with power set to 0.8. All experiments are trained on eight NVIDIA 4090 GPUs with a batch size of one per GPU.

\begin{table}[]
\centering
\setlength{\tabcolsep}{1.5mm} %
\begin{tabular}{l|c|c|c} 
\toprule
Method & Vehicle & Pedestrian & Cyclist \\
\midrule
& \multicolumn{3}{c}{MOTA $\uparrow$ / IDS \%$\downarrow$} \\
\cmidrule(lr){2-4}
CenterPoint & 55.1 / 0.26 & 54.9 / 1.13 & 57.4 / 0.83  \\
SpOT & 55.7 / 0.18 & 60.5 / 0.56 & -- / -- \\ 
TrajectoryFormer & 59.7 / 0.19 & 61.0 / 0.37 & 60.6 / 0.70 \\ 
\midrule
\textbf{DSC-Track(ours)} & \textbf{60.5 / 0.11} & \textbf{61.1 / 0.27} & \textbf{60.9 / 0.18} \\
\bottomrule
\end{tabular}
\caption{Tracking results (MOTA / IDS\%) on Waymo validation split.}
\label{tab:waymo_val_combined}
\end{table}

\subsection{Main Results}

To validate our approach, we compare DSC-Track against state-of-the-art methods on the nuScenes and Waymo benchmarks. On nuScenes, we compare against two categories of baselines: methods leveraging additional \textbf{3D appearance} cues like ShaSTA~\cite{sadjadpour2023shasta} and NEBP~\cite{liang2023neural}, and those without, including strong competitors like 3DMOTFormer~\cite{Ding2023MOTFormer}, PolarMOT~\cite{kim2022polarmot}, ImmortalTracker~\cite{wang2021immortal}, SimpleTrack~\cite{pang2022simpletrack}, and CBMOT~\cite{benbarka2021score}. On Waymo, we compare against recent strong methods such as SpOT~\cite{stearns2022spot} and TrajectoryFormer~\cite{chen2023trajectoryformer}. For fair comparison, all methods are built upon detections from the widely-used CenterPoint~\cite{yin2021center} detector.

\paragraph{Results on the nuScenes Test Set.}
Our primary results on the challenging nuScenes test set (Table~\ref{tab:nuscenes_test}) show that DSC-Track establishes a new state-of-the-art with 70.3\% AMOTA. Notably, this is achieved without relying on any appearance features, yet our method surpasses the best appearance-based method, ShaSTA, by 0.7\% and outperforms 3DMOTFormer by a significant 2.1\% in AMOTA. This superiority extends to other crucial metrics; it achieves the best AMOTP (0.476), indicating superior localization precision, and its IDS score marks a dramatic 31.3\% reduction in identity switches compared to 3DMOTFormer. The class-level analysis further underscores its robustness, as it secures the highest AMOTA for 5 out of 7 categories, including the critical car, pedestrian, and bus classes.

\paragraph{Validation on nuScenes and Generalization to Waymo.}
This strong performance is mirrored on the nuScenes validation set (Table~\ref{tab:nuscenes_val}), where DSC-Track again sets a new state-of-the-art with 73.2\% AMOTA, surpassing 3DMOTFormer by 2.0\%. To evaluate generalizability, we also test on the Waymo Open Dataset (Table~\ref{tab:waymo_val_combined}), where our method excels at tracking stability, securing the best or tied-for-best IDS scores across all major categories: Vehicle, Pedestrian, and Cyclist.

\paragraph{Runtime Speed.}
Our approach runs at a rate of 26.77 FPS on a single NVIDIA 4090 GPU, which is efficient for real-time multi-object tracking.

\subsection{Ablation Study and Analysis}
\label{sec:ablation}

\paragraph{Analysis of Different Components.}
Our ablation study in Table~\ref{tab:component_ablation} confirms the indispensability of each module. Removing the Temporal Encoder is most detrimental, causing a 5.7\% AMOTA drop and a massive surge in IDS, underscoring the necessity of historical context. Ablating the Cue-Consistent Attention leads to a 2.7\% AMOTA loss and a threefold increase in IDS, validating our explicit matching mechanism. Finally, removing the Geometric Encoder results in a 1.4\% AMOTA drop, proving the importance of spatial relationship encoding. These results affirm our model's performance stems from the synergy of its components.

\begin{table}[tb]
\centering
\setlength{\tabcolsep}{0.75mm}
\begin{tabular}{c|cccc}
\toprule
Method & AMOTA & AMOTP & MOTA & IDS \\
\midrule
w/o Geometric Encoder & 71.8 & 0.510 & 61.3 & 450 \\
w/o Temporal Encoder & 67.5 & 0.580 & 57.0 & 1015 \\ 
w/o Cue-Consistent Attn & 70.5 & 0.505 & 60.1 & 950 \\
\midrule
\textbf{Ours} & \textbf{73.2} & \textbf{0.498} & \textbf{62.5} & \textbf{298} \\
\bottomrule
\end{tabular}
\caption{Analysis of different components using the nuScenes validation set.}
\label{tab:component_ablation}
\end{table}

\paragraph{Analysis of Initial Feature Embedding.}
As shown in Table~\ref{tab:ppf}, our initial feature embedding design relies on the synergy between relative geometry (PPF) and contextual features. Removing the PPF-based geometry (w/o PPF) impairs pose-sensitivity, resulting in a 1.1\% AMOTA drop and a 29\% IDS increase. Conversely, removing contextual features (w/o Contextual) is even more detrimental, causing a 1.2\% AMOTA loss and a 54\% surge in IDS due to the lack of global scene context. These results confirm that both information sources are complementary and essential for constructing robust discriminative features.

\begin{table}[tb]
\centering
\setlength{\tabcolsep}{1mm}
\begin{tabular}{c|cccc}
\toprule
Method & AMOTA & AMOTP & MOTA & IDS \\
\midrule
w/o PPF & 72.1 & 0.510 & 61.9 & 385 \\
w/o Contextual & 72.0 & 0.512 & 61.7 & 460 \\
\midrule
\textbf{Ours} & \textbf{73.2} & \textbf{0.498} & \textbf{62.5} & \textbf{298} \\
\bottomrule
\end{tabular}
\caption{Ablation study on the the features of geometric encoders.}
\label{tab:ppf}
\end{table}

\paragraph{Analysis of Cue-Consistent Cross-Attention.}
We compare our Cue-Consistent Attention with two alternatives in Table~\ref{tab:interaction}. A Vanilla Cross-Attention struggles without explicit guidance, leading to a 0.4\% AMOTA drop and a 31\% IDS increase. A simpler Max Pooling aggregation performs worse, with AMOTA decreasing by 0.9\%, as it discards contextual information. This superior performance validates our design: by using pre-computed self-attention scores to guide the interaction, it robustly aggregates consistent features for more accurate associations.

\begin{table}[tb]
\centering
\setlength{\tabcolsep}{1mm}
\begin{tabular}{c|cccc}
\toprule
Method & AMOTA & AMOTP & MOTA & IDS \\
\midrule
Max Pooling & 72.3 & 0.515 & 61.4 & 480 \\
Vanilla Cross-Attn & 72.8 & 0.508 & 61.8 & 390 \\
\midrule
\textbf{Ours} & \textbf{73.2} & \textbf{0.498} & \textbf{62.5} & \textbf{298} \\
\bottomrule
\end{tabular}
\caption{Ablation study on different interaction mechanisms for cue-consistency.}
\label{tab:interaction}
\end{table}

\begin{figure}[tb]
\centering 
\includegraphics[width=0.48\textwidth]{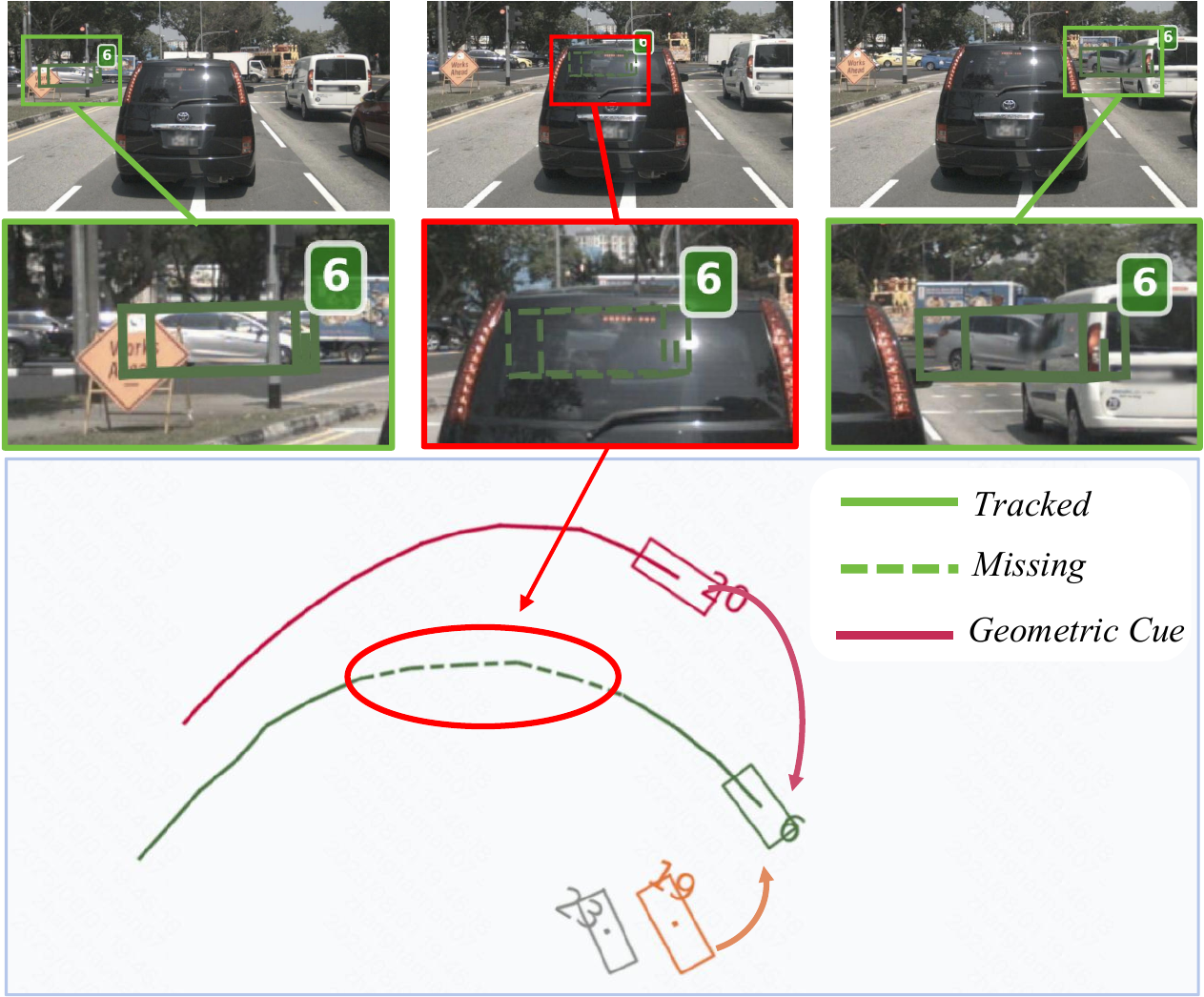}
\caption{Qualitative results of DSC-Track on the nuScenes validation set. Our tracker successfully handles a long-term occlusion during a turn by leveraging stable geometric cues from the environment (e.g., the roadside), correctly re-identifying the target (ID 6) upon reappearance.}
\label{fig/qua_vis} 
\end{figure}

\subsection{Qualitative Results}
Figure~\ref{fig/qua_vis} showcases the robustness of DSC-Track in a challenging scenario with long-term occlusion during a turn. While conventional trackers often fail due to non-linear motion and prolonged target absence, our method successfully bridges the gap. As shown, despite the target vehicle (ID 6) being fully occluded, our approach leverages stable geometric cues from the environment (e.g., the roadside relationship). This allows our tracker to maintain the target's state during occlusion and robustly re-identify it upon reappearance, ensuring trajectory integrity.

\section{Conclusion}
In this paper, we propose \textbf{\name{}} to address data association in 3D MOT. Based on cue-consistency, our method leverages rotation-invariant Point Pair Features (PPF) and a cue-consistent attention strategy to match trajectories and detections by their neighborhood structures, rather than individual features. Experiments show \name{} achieves state-of-the-art performance and significantly reduces identity switches in complex scenarios. Our work validates modeling higher-order relational consistency over object-level cues as a promising direction for robust tracking.

\section*{Acknowledgments}
This research was supported by The National Nature Science Foundation of China (Grant Nos: 62402417, 62273301, 62273302), in part by "Pioneer" and "Leading Goose" R\&D Program of Zhejiang (Grant No. 2025C02026), in part by the Key R\&D Program of Ningbo (Grant Nos: 2024Z115, 2025Z035), in part by Yongjiang Talent Introduction Programme (Grant No: 2023A-197-G).

\bibliography{aaai2026}

\end{document}